# Rational Nonmonotonic Reasoning

CARL M. KADIE (KADIE@M.CS.UIUC.EDU)

Department of Computer Science, University of Illinois,
Urbana, IL 61801 U.S.A.

**Abstract.**

Nonmonotonic reasoning is a pattern of reasoning that allows an agent to make and retract (tentative) conclusions from inconclusive evidence. This paper gives a possible-worlds interpretation of the nonmonotonic reasoning problem based on standard decision theory and the emerging probability logic. The system's central principle is that a tentative conclusion is a decision to make a bet, not an assertion of fact. The system is rational, and as sound as the proof theory of its underlying probability logic.

## 1 Introduction and Background

The ability to make a tentative conclusion from inconclusive evidence and then retract that conclusion if discrediting evidence is later acquired is characteristic of intelligence. Many theories of nonmonotonic reasoning, default reasoning, and inductive logic try to model this behavior. But, because these theories do not fully consider the consequences and context of adopting a conclusion, they can behave irrationally. As the expression "Don't bet your life on it" illustrates, hasty conclusions can be disastrous.

Decision theory and the emerging probability logic offer a straightforward possible-worlds interpretation of idealized nonmonotonic reasoning that adapts to the context of the reasoning problem.

### 1.1 The Bird Example

The bird example demonstrates nonmonotonic reasoning.

The nonmonotonic reasoner is told: "Tweety is a bird."

It is asked: "Can Tweety fly?"

According to the literature, it answers: "Yes, Tweety can fly."

Now it is given another fact: "Tweety is a Penguin."

Now, how should it answer: "Can Tweety Fly?"

According to the literature, it answers: "No, Tweety can't fly."

The addition of new facts caused the nonmonotonic reasoner to retract a conclusion. In conventional logic, new facts never cause an agent to retract a conclusion; thus, conclusions grow monotonically with information. Consequently, conventional logic must be more conservative in making conclusions than nonmonotonic reasoning.

### 1.2 Current Nonmonotonic Reasoning Systems

Logic and probability are two popular approaches to nonmonotonic reasoning.

### 1.2.1 Logic-Based Systems

Conceptually, a logic-based nonmonotonic reasoner makes its conclusions in two steps. First, from its defaults and knowledge it creates deductively-closed belief sets called extensions (Ginsberg, 1987). In the Tweety example, with default $bird(x) \rightarrow fly(x)$ and the knowledge that $bird(Tweety)$ and $penguin(x) \rightarrow \neg fly(x)$, it creates the single extension $\{bird(Tweety), penguin(x) \rightarrow \neg fly(x), fly(Tweety)\}$. The second part of making conclusions is the application of an acceptance rule. The rule is a procedure that tells the nonmonotonic reasoner what conclusions to draw from the extensions. When a nonmonotonic system creates a single extension, the acceptance rule is simple: accept (or conclude) a sentence $S$ if it is in the (deductive closure of the) extension.

Continuing the example, if the system is told that *penguin(Tweety)*, the default rule cannot fire because its consequent contradicts the fact $\neg fly(Tweety)$ (from *penguin(Tweety)* and $penguin(x) \rightarrow \neg fly(x)$). Thus, it now concludes that Tweety cannot fly because it now has the single extension: $\{bird(Tweety), penguin(x) \rightarrow \neg fly(x), \neg fly(Tweety)\}$.

The acceptance rules are more interesting when defaults conflict. For example, suppose the reasoner knows that *Quaker(Nixon)* and *Republican(Nixon)* and that it has defaults $Quaker(x) \rightarrow pacifist(x)$ and $Republican(x) \rightarrow \neg pacifist(x)$ (Reiter, Criscuolo, 1981). Now, rather than one extension there are two (Moore, 1985):

$$\{Quaker(Nixon), Republican(Nixon), pacifist(Nixon)\}$$

and

$$\{Quaker(Nixon), Republican(Nixon), \neg pacifist(Nixon)\}.$$

Reiter's system *accepts* a sentence (tentatively concludes that it is so) if the sentence is true in *any* extension (Reiter, 1980). Consequently, it accepts the sentence *pacifist(Nixon)* and the sentence $\neg pacifist(Nixon)$. McDermott (1982) calls this acceptance rule *brave*. Circumscription offers a different acceptance rule: it accepts a sentence if the sentence is true in *every* extension. Thus, circumscription makes no conclusion about *pacifist(Nixon)* and no conclusion about $\neg pacifist(Nixon)$. McDermott calls this acceptance rule *cautious*.

These two acceptance rules will be called *modal*. If the set of extensions is treated as the set of possible worlds in an S5 modal structure, Reiter's acceptance rule corresponds closely to the model operator $\diamond$ (possibility) and circumscription's rule corresponds closely to the $\square$ (necessity) operator (Dowty, Wall, and Peters, 1981).

Many systems forgo the modal approach by evaluating all sentences in a particular extension. The extension can be chosen arbitrarily (Doyle, 1983), according to some hierarchy of defaults (or priorities of circumscription (McCarthy, 1980; Lifschitz, 1987)), by choosing the most probable extension (Pearl, 1987), or by some other method.

In these pick-an-extension systems the truth value of a complex sentence is a function of the truth values of the literals that make up that sentence. This is not true in the modal systems. For example, Reiter's modal system does not accept the sentence $pacifist(Nixon) \land \neg pacifist(Nixon)$, but it does accept the literal *pacifist(Nixon)* and the literal $\neg pacifist(Nixon)$. Likewise, circumscription accepts $pacifist(Nixon) \lor \neg pacifist(Nixon)$, but not *pacifist(Nixon)*, and not $\neg pacifist(Nixon)$.

### 1.2.2 Probability-Based Systems

Like logic-based systems, probability-based nonmonotonic reasoning systems (conceptually) make conclusions in two steps. First, they assign some degree of belief to a sentence. The degree might be a probability measure, a certainty factor, a Zadeh possibility degree, or a Dempster-Shafer interval. Second, using the degree of belief, an acceptance rule decides whether to make the sentence a conclusion. (Some AI systems (Rich, 1983; Ginsberg, 1984; Farreny & Prade, 1986; Dubois & Prade, 1988) ignore the acceptance step). The most obvious acceptance rule is: Accept $S$ if, and only if,

$$P(S) > 1 - \varepsilon, \text{ for some small } \varepsilon.$$

$P(S)$ is the probability of sentence $S$. The rule says: if a sentence is certain enough, say 99.99% certain, conclude that it is so.

### 1.3 Scope

The body of this paper concerns the desired properties of a nonmonotonic reasoning system and the specification of such a system. Before starting, it is appropriate to discuss what will and will not be done.

First, this paper will take no stand in the long-running debate as to the practicality of probabilities:

> *The information necessary to assign numerical probabilities is not ordinarily available. Therefore, a formalism that required numerical probabilities would be epistemologically inadequate. (McCarthy & Hayes, 1969)*

And on the other hand:

> *... many extensions of logic, such as 'default logic' are better understood in a probabilistic framework. (Cheeseman, 1985)*

Second, this paper will present a formal account of nonmonotonic reasoning. Its approach may superficially appear less formal than approaches that invent a new formal system and provide proofs of its properties. The approach taken here *unifies* (Ginsberg, 1987) nonmonotonic reasoning with a previously developed formal system. This approach may legitimately rely on others to develop the formal properties of the formal system.

Third, this paper will not prove that this nonmonotonic system is correct or optimal because there is no standard to which it can be compared. All that will be argued is that the system offers useful insight into interesting problems, that it make only reasonable assumptions, and that it is internally consistent.

Now that the general scope of the desired nonmonotonic system is delimited, its particular properties are specified.

## 2 Desired Properties

The desired properties of *soundness* and *rationality* are described and justified in turn.

### 2.1 Soundness

One test of a nonmonotonic system's soundness is Kyburg's lottery paradox (Kyburg, 1970). The paradox shows that nonmonotonic systems that allow chaining (that is, *and introduction*, *modus ponens*, and so on) across extensions (for example (Ginsberg, 1985; Yager, 1987)) are unsound; that is, they allow inconsistent conclusions. Kyburg's example was this: suppose there is a lottery with 10000 tickets. A reasoning system knows that exactly one ticket will win. Suppose it has a default that says that (unless it knows to the contrary) it should assume that *loser(ticket_#1)*. Similarly, suppose it has a default that says *loser(ticket_#2)* and so on for all the tickets. These defaults and the fact that exactly one ticket will win lead to 10000 extensions. In each extension a different ticket wins.

If the system is brave with respect to the literals, its set of conclusions will be *loser(ticket_#1), loser(ticket_#2), ..., loser(ticket_#10000)*. And if chaining is allowed across extensions, then by *and introduction* it will conclude that *loser(ticket_#1)* ∧ *loser(ticket_#2)* ∧ ... ∧ *loser(ticket_#10000)*. That is, it will conclude that no ticket will win. This conclusion contradicts its knowledge that one ticket will win.

Two alternatives to chaining across extensions were mentioned earlier: pick-an-extension systems, which chain within a single extension, and modal systems, which use many extensions, but allow no chaining. As a third alternative, a system may be particularly cautious about accepting literals; it may accept only the deductive closure of the literals true in every extension. In the lottery example, because no literal is true in every extension, such a system would accept nothing; thus it avoids all the problems (and benefits) of nonmonotonic reasoning.

### 2.2 Rationality

The other property to consider is rationality. Rationality is best understood with an example.

#### 2.2.1 Russian Roulette Example

Consider this problem in nonmonotonic reasoning:

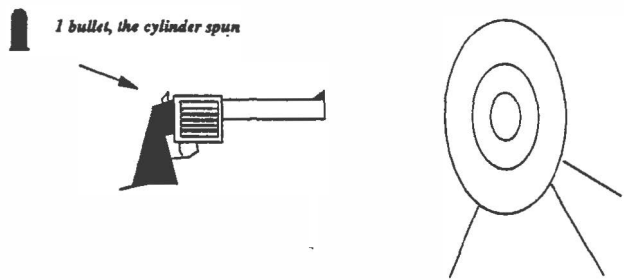

A revolver is loaded with 1 bullet (it has 5 empty chambers), and the cylinder is spun.
With these stakes:

> If correct, the system wins $1.
> If wrong, the system loses $1.

would a nonmonotonic system take the bet that the gun will *not* fire? The answer is yes; most systems would take the bet, because the gun is unlikely to fire.

Now consider a second scenario:

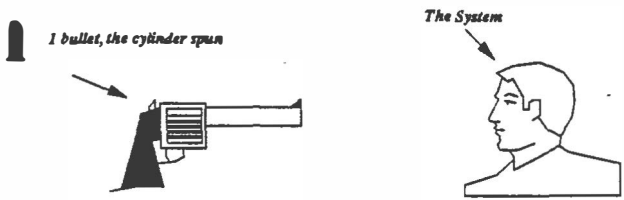

Again the revolver is loaded with exactly 1 bullet and the cylinder is spun.
With these new stakes:

> If correct, the system wins $1.
> If wrong, the system loses its life.

would a nonmonotonic system take the bet that the gun will not fire?

Most nonmonotonic systems would take the bet -- because, the gun is still unlikely to fire.

Obviously something is wrong with this reasoning; in these two scenarios the uncertainty is the same, yet it is not *rational*[1] to draw the same conclusion.

How is the Russian roulette problem solved? Decision theory gives the answer: compare the probability of the sentence to the breakeven probability determined by the payoff. In the first scenario, the breakeven probability is $1/(1 + 1) = 0.5$, and because $P(gun\_will\_not\_fire \mid 1\_bullet\_and\_spun)$ is greater than 0.5, the system should conclude that the gun will not fire. In the second scenario the breakeven probability is

$$\lim_{i \to \infty} \frac{i}{i + 1} = 1.$$

The system should ignore a better-than-even probability and refuse to bet its life on the proposition that the gun will not fire.

---

[1] Though the exact meaning of the term *rational* is the subject of debate, most authors agree that a rational agent maximizes its expected utility (Savage, 1954; Ferguson, 1967) and avoids Dutch books (Kyburg & Smokler 1964).

### 2.2.2 The Need for Payoff and Probability

The example shows that rational nonmonotonic systems must consider payoff and act accordingly. Systems that fail to consider payoff will in general miss opportunities (the first scenario) or suffer disastrous consequences (the second). How should payoff be measured? If it is assumed there is a total preference ordering among payoffs and that the preference ordering of bets is based on expected payoff, *utility theory* is the answer (Ferguson, 1967; Savage, 1954). Furthermore, if these axioms of utility theory are accepted, it follows that a rational nonmonotonic system must be probabilistic, for without probability, rational choices cannot be determined.

Decision theory allows us to have a nonmonotonic reasoner that automatically adjusts its degree of valor. Depending on the circumstances, it can be cautious like a circumscription system, brave like Reiter's system, or somewhere in between.

Consideration of payoff is opposed by some. Pearl (1987) asserts that in day-to-day life the overhead of considering payoff overwhelms the benefit of such deliberations. Others disagree (Levi 1980). In an idealized system there is no debate -- computation is free and optimal performance is expected -- so payoff must be considered.

## 3 Specification

Given these desired properties, a description of a rational nonmonotonic system is now straightforward. The nonmonotonic system is based on decision theory and probability. It is modal and does not allow chaining.

### 3.1 Probability

The first step is the specification of a language, model theory, and semantics. One possibility is Nilsson's probabilistic logic (Nilsson, 1986).

The intuition behind probabilistic logic is that a particular sentence in a particular world/extension is either true or false. But, because agents do not know which world they are in, they must look at the set of worlds consistent with their knowledge. Agents find the probability of a sentence by summing the probability of each world in which the sentence is true.

If the set of literals is finite, the set of all possible worlds and their probabilities is expressible as a finite joint probability density. For example the prior density for the bird example might be

|               | Can Fly | Can't Fly |             |
|---------------|---------|-----------|-------------|
| Not a Penguin | 0.002   | 0.888     | Not a Bird  |
| A Penguin     | 0       | 0         |             |
| Not a Penguin | 0.090   | 0.010     | A Bird      |
| A Penguin     | 0       | 0.010     |             |

Although probabilistic logic gives the desired model theory, it lacks expressiveness, so Frisch & Haddawy's modal probability logic (Frisch & Haddawy, forthcoming), an extended probability logic, is used, making it possible to say things such as

$$P(gun\_will\_not\_fire \mid 1\_bullet\_and\_spun) > 0.5$$

in the object language.

Because the nonmonotonic reasoning system is embedded in a probability logic, it is as sound (and complete) as that probability logic. A proof theory for the propositional probability logic is trivial: simply enumerate the finite possible-worlds and add their probabilities. This proof theory is sufficient for the examples considered in this paper. A proof theory for predicate probability logic is still emerging.

## 3.2 Acceptance

The next step is the specification of the acceptance rule. A sentence $S$ with respect to some evidence $e$ and a breakeven probability $b$ is accepted exactly if

$$P(S \mid e) > b.$$

This rule comes from decision theory.

The key to this system, therefore, is that a tentative conclusion is an assertion about the desirability of a bet, not a direct assertion about a sentence. For example, where other systems might conclude $fly(Tweety)$, this system concludes $P(fly(Tweety) \mid bird(Tweety)) > 0.6$. If later it learns that $penguin(Tweety)$, it concludes $\neg P(fly(Tweety) \mid bird(Tweety) \& penguin(Tweety)) > 0.6$.

## 3.3 Chaining

The final part of the specification is a prohibition against chaining conclusions. The system is not allowed to take

$$P(loser(ticket\_\#1) \mid 10000\_ticket\_lottery) > \alpha$$

and

$$P(loser(ticket\_\#2) \mid 10000\_ticket\_lottery) > \alpha$$

and deduce

$$P(loser(ticket\_\#1) \wedge loser(ticket\_\#2) \mid 10000\_ticket\_lottery) > \alpha.$$

The prohibition is justified because $loser(ticket\_\#1)$ and $loser(ticket\_\#2)$ might seldom be true in the same world, and so, any chaining crosses extensions/worlds.

Actually, no explicit prohibition is needed because the semantics (from probability logic) already prohibit chaining across worlds. Also, because sentences such as

$$P(loser(ticket\_\#1) \wedge loser(ticket\_\#2) \mid 10000\_ticket\_lottery) > \alpha,$$

can be evaluated directly, no chaining needed.

## 3.4 Tweety Revisited

Here is how this system renders the bird example. Suppose 1) that the prior joint probability density is

|               | Can Fly | Can't Fly |            |
| ------------- | ------- | --------- | ---------- |
| Not a Penguin | 0.002   | 0.888     | Not a Bird |
| A Penguin     | 0       | 0         |            |
| Not a Penguin | 0.090   | 0.010     | A Bird     |
| A Penguin     | 0       | 0.010     |            |

and 2) that the payoff is

| The system says $can\_fly(Tweety)$. If right, it wins $1.00. If wrong, it loses $1.50. | The system says $\neg can\_fly(Tweety)$. If right, it wins $1.50. If wrong, it loses $1.00. |
| --- | --- |

(the breakeven probabilities are 0.6 and 0.4, respectively.

and 3) that the reasoner is told $bird(Tweety)$. The probabilities calculated from the density are:

$P(can\_fly(Tweety) \mid bird(Tweety) =$

$$\frac{0 + 0.09}{0 + 0.09 + 0.01 + 0.01} \approx 0.82$$

and

$P(\neg can\_fly(Tweety) \mid bird(Tweety)) =$

$$\frac{0.01 + 0.01}{0 + 0.09 + 0.01 + 0.01} \approx 0.18.$$

Because 0.82 is greater that 0.6, the system concludes
$$P(can\_fly(Tweety) \mid bird(Tweety)) > 0.6;$$
and because 0.18 is not greater than 0.4 it also concludes
$$\neg P(\neg can\_fly(Tweety) \mid bird(Tweety)) > 0.4.$$
Or in English: given that Tweety is a bird, it is rational to bet Tweety can fly (and not to bet that Tweety cannot fly). Now suppose that the system is told that Tweety is a penguin. From the probability density the system computes that

$P(can\_fly(Tweety) \mid bird(Tweety) \wedge penguin(Tweety)) =$

$$\frac{0}{0.01 + 0} = 0$$

and

$P(\neg can\_fly(Tweety) \mid bird(Tweety) \wedge penguin(Tweety)) =$

$$\frac{0.01}{0.01 + 0} = 1.$$

It concludes
$$\neg P(can\_fly(Tweety) \mid bird(Tweety) \wedge penguin(Tweety)) > 0.6$$
and
$$P(\neg can\_fly(Tweety) \mid bird(Tweety) \wedge penguin(Tweety)) > 0.4.$$
That is, given what it knows, it bets that Tweety cannot fly (and it does not bet that Tweety can fly).

So the system handles the bird example properly.

## 4 Conclusion

A nonmonotonic reasoning system should be sound and rational. The Russian roulette problem shows the dangers of irrational reasoning. A system that does not evaluate probabilities and consider payoff may miss opportunities and stumble into disasters. With decision theory, a system is automatically as courageous as circumstances warrant. In the extremes, it can be as cautious as a circumscription system or as brave as Reiter's system.

Probability logic gives a decision-theory-based nonmonotonic reasoning system a rich possible-worlds semantics. The form of the conclusions is

$$P(S \mid e) > b$$

or

$$\neg P(S \mid e) > b,$$

where $S$ is the sentence in question, $e$ is the current knowledge (evidence), and $b$ is the breakeven probability determined from the payoff.

The system has all the desired properties (if the proof theory of its probability logic is sound), and so provides insight into nonmonotonic reasoning, an important facet of intelligent behavior.

### Acknowledgments

This work was supported in part by the Office of Naval Research under grant N00014-88-K-0124. Thanks also to Larry Rendell, Peter Cheeseman, and Marianne Winslett for their valuable comments on an early version of this paper.

# REFERENCES

Cheeseman, P. (1985). In defense of probability. In *Proc. IJCAI-85*.

Dowty, D.R., Wall, R.E., & Peters, S. (1981). *Introduction to Montague Semantics*. Dordrecht, Holland: Deidel Publishing.

Doyle, J. (1983). A truth maintenance system. *Artificial Intelligence, 12*, 231-272.

Dubois, D. & Prade, H. (1988). Default reasoning and possibility theory. *Artificial Intelligence, 35*, 243-257.

Farreny, H. & Prade, H. (1986). Default and inexact reasoning with possibility degrees. *Tran. on Systems, Man, and Cybernetics, 16*, no. 2., 270-276.

Ferguson, T.S. (1967). *Mathematical Statistics: A Decision Theoretic Approach*. New York: Academic Press.

Frisch, A.M. & Haddawy, P. (forthcoming). Probabilistic as a modal operator. *Proc. of the Fourth Workshop on Uncertainty in Artificial Intelligence*.

Ginsberg, M.L. (1984). Non-monotonic reasoning using Dempster's rule. In *Proc. AAAI-84* (pp. 126-129).

Ginsberg, M.L. (1985). Does probability have a place in non-monotonic reasoning? In *Proc. IJCAI-85* (pp. 107-110).

Ginsberg, M.L. (1987). Introduction of *Readings in Nonmonotonic Reasoning*, Los Altos, CA: Morgan Kaufmann Publishers.

Kyburg, H.E. & Smokler, H.E. (1964). Introduction of *Studies in Subjective Probability*. New York: John Wiley & Sons.

Kyburg, H.E. (1970). Conjuctivitis. In M. Swain (Ed.), *Induction, Acceptance, and Rational Belief*. Reidel.

Levi, I. (1980). *The Enterprise of Knowledge*. Cambridge, MA: MIT Press.

Lifschitz, V. (1987). *Pointwise Circumscription* (Technical Report). Stanford, CA: Stanford, University. Reprinted in M.L. Ginsberg (Ed.), *Readings in Nonmonotonic Reasoning*. Los Altos, CA: Morgan Kaufmann Publishers.

McCarthy, J. & Hayes, P. (1969). Some philosophical problems from the standpoint of artificial intelligence. *Machine Intelligence 4*. Reprinted in B. Webber & N. Nilsson (Ed.), *Readings in Artificial Intelligence*. Tioga, 1981.

McCarthy, J. (1980). Circumscription -- A form of non-monotonic reasoning. *Artificial Intelligence, 13*, 27-39.

McDermott, D. (1982). Non-monotonic logic II. *Journal ACM, 29*, 33-57.

Moore, R.C. (1985). Semantical considerations on nonmonotonic logic. *Artificial Intelligence, 25*, 75-94.

Nilsson, N. (1986). Probabilistic logic. *Artificial Intelligence, 28*.

Pearl, J. (1987). Distributed revision of composite beliefs. *Artificial Intelligence, 33*, 173-215.

Reiter, R. & Criscuolo, G. (1981). On interacting defaults. *Proc. IJCAI-81* (pp. 270-276).

Reiter, R. (1980). A Logic for Default Reasoning. *Artificial Intelligence, 13*, 81-132.

Rich, E. (1983). Default reasoning as likelihood reasoning. *Proc. AAAI-83* (pp. 348-351).

Savage, L.J. (1954). *The Foundations of Statistics*. John Wiley & Sons.

Yager, R.R. (1987). Using approximate reasoning to represent default knowledge. *Artificial Intelligence, 31*, 99-112.